\documentclass[lettersize,journal]{IEEEtran}
\usepackage{amsmath,amsfonts}
\usepackage{algorithmic}
\usepackage{array}
\usepackage[caption=false,font=footnotesize,labelfont=rm,textfont=rm]{subfig}
\usepackage{textcomp}
\usepackage{stfloats}
\usepackage{url}
\usepackage{verbatim}
\usepackage{graphicx}
\usepackage{cite}
\usepackage{booktabs}
\usepackage{multirow}
\usepackage{colortbl}
\usepackage{color}
\usepackage{tabularx}
\usepackage[dvipsnames]{xcolor}
\usepackage[ruled, linesnumbered]{algorithm2e}
\usepackage[colorlinks, linkcolor=blue, anchorcolor=blue, citecolor=blue]{hyperref}
\begin{document}

\title{A Benchmark Dataset for MLLM-Generated Image Detection: GPT Image2 \& Nano Banana2}

\author{Zirui Zhang, Yinbo Yu, Donghai Guan, Chunwei Tian, Daoqiang Zhang,~\IEEEmembership{Senior Member, IEEE,} and Qi Zhu

\thanks{Zirui Zhang and Donghai Guan are with the College of Computer Science and Technology, Nanjing University of Aeronautics and Astronautics, Nanjing 210016, China (e-mails: zhangzirui@nuaa.edu.cn, dhguan@nuaa.edu.cn).}
\thanks{Yinbo Yu, Daoqiang Zhang, and Qi Zhu are with the College of Artificial Intelligence, Nanjing University of Aeronautics and Astronautics, Nanjing 210016, China (e-mails: yinboyu@nuaa.edu.cn, dqzhang@nuaa.edu.cn, zhuqinuaa@163.com).}
\thanks{Chunwei Tian is with the School of Computer Science and Technology, Harbin Institute of Technology, Harbin 150001, China (e-mail: chunweitian@163.com).}
\thanks{Corresponding authors: Donghai Guan, and Qi Zhu.}
}



\maketitle

\begin{abstract}
The realism of images generated by multimodal large language models (MLLMs), such as GPT Image2 and Nano Banana2, has improved rapidly in recent years. Compared with early generative models, current models have made clear progress in text rendering. They can produce high-quality images that closely resemble real-world application scenarios. The enhanced generation capabilities of current MLLMs pose increasingly severe challenges to AI-generated image detection. Detection is no longer limited to identifying obvious artifacts left by early generators. Instead, it requires systematic and realistic benchmarks for the new generation of generated content. However, most existing benchmarks are still built around early generative models and cannot fully evaluate the forensic challenges introduced by high-quality and multi-form generated images. To address this gap, this paper constructs a benchmark dataset for detecting images generated by MLLMs. The benchmark covers several realistic application scenarios and adopts three generation protocols to simulate direct generation, reference-based reconstruction, and local editing. Based on this benchmark, we evaluate detector degradation from traditional scenarios to MLLM-generated images and analyze false positive rates and false negative rates across three sample types, revealing the failure modes of existing methods. We further propose a \textbf{s}tructural-\textbf{a}rtifact-\textbf{p}rior-guided \textbf{d}ual-\textbf{s}tream \textbf{p}rompt framework (SAP-DSP) as a strong baseline. SAP-DSP uses dual-stream prompt learning and structure-aware routing fusion to improve representation learning. Extensive experiments show that the proposed benchmark exposes the performance degradation of existing detectors on high-quality generated images, while SAP-DSP achieves more stable detection results on this benchmark. Our code and dataset are publicly available at https://github.com/xbrainnet/SAP-DSP.
\end{abstract}

% \begin{IEEEkeywords}
% Biometric, Contactless Palmprint, Generative Adversarial Network, Memory Evolution.
% \end{IEEEkeywords}

\section{Introduction}
\IEEEPARstart{R}{ecent} multimodal large language models (MLLMs), such as GPT Image2 and Nano Banana2, are reshaping AI image synthesis by shifting generated images from generic natural image synthesis toward more realistic visual content. Compared with earlier generative models~\cite{ref1, ref2, ref3}, current models can not only generate high-quality facial and scene images, but also follow text prompts to produce visual content that closely matches real-world scenarios. The boundary between real and generated images is becoming increasingly blurred. Therefore, AI-generated image detection must address more realistic, diverse, and challenging images.

\begin{figure}[t]
  \centering
  \includegraphics[width=\linewidth]{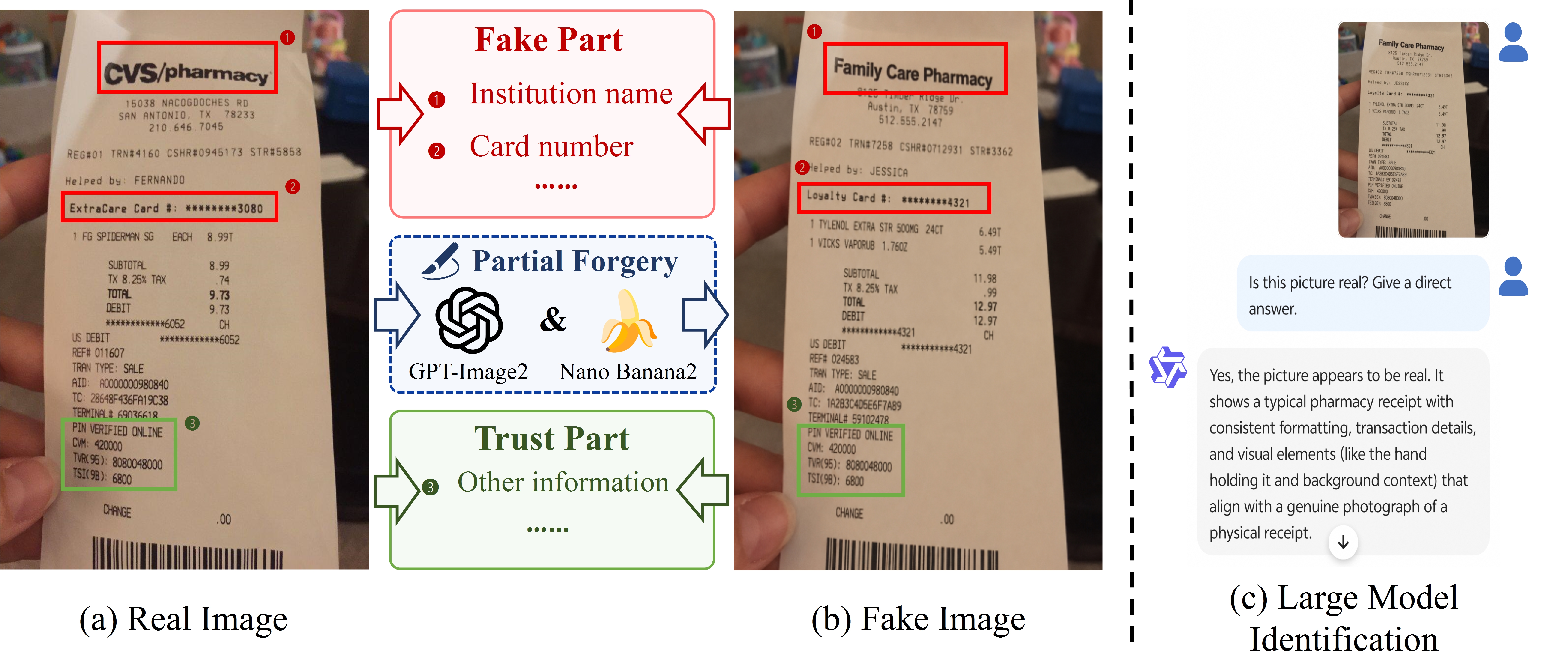}
  \caption{An example of local image forgery}
  \label{fig1}
\end{figure}

Existing studies on AI-generated image detection have made important progress. Numerous methods mine generation traces from texture anomalies or local noise patterns and achieve promising performance on public benchmarks such as UFD~\cite{ref4} and GenImage~\cite{ref5}. However, these benchmarks are mostly built upon traditional GANs~\cite{ref6, ref7} or diffusion models~\cite{ref8, ref9}, and therefore cannot fully reflect the new challenges introduced by GPT Image2 and Nano Banana2. On the one hand, artifacts in current generated images are much weaker, making low-level texture cues and model-specific traces less reliable. On the other hand, generated content in real-world applications is no longer limited to natural images. It also appears in web screenshots, receipts, documents, and locally edited images. Such samples contain natural textures, text boundaries, and cross-region semantic relations, which further increases the complexity of detection. Figure~\ref{fig1} shows an example in which a large image generation model is used to locally modify key information in an image. The resulting fake sample is highly realistic. When this sample is submitted to another multimodal large model for authenticity judgment, the model fails to identify it as fake.

To systematically evaluate existing detectors on images generated by MLLMs, we construct a benchmark based on GPT Image2 and Nano Banana2. The benchmark covers three types of visual content: texture-dominated images, structure-dominated images, and hybrid-dominated images. These categories simulate natural images, structured documents, and multi-region mixed content in real-world applications. During dataset construction, we introduce three generation protocols, including direct generation, reference-based reconstruction, and local editing. These protocols increase the diversity of sample sources and generation traces. The dataset also contains text-image content with different layouts, enabling the evaluation of detector robustness in complex real-world scenarios.

Based on this benchmark, we first conduct a systematic evaluation of representative AI-generated image detectors from both macro and micro perspectives. At the macro level, we compare multiple detector performance metrics, such as accuracy, on both public benchmarks and the self-built benchmark. This comparison shows how performance degrades when detectors are transferred from traditional settings to images generated by GPT Image2 and Nano Banana2. At the micro level, we compute false positive rates and false negative rates on Texture, Structure, and Hybrid samples, revealing how existing detectors fail on both real and generated images.

We further propose a \textbf{s}tructural-\textbf{a}rtifact-\textbf{p}rior-guided \textbf{d}ual-\textbf{s}tream \textbf{p}rompt framework (SAP-DSP) as a strong structure-aware baseline for this task. From the perspective of forensic evidence, SAP-DSP decomposes MLLM-generated image detection into texture evidence modeling and structural evidence modeling. Texture evidence mainly reflects traditional generation traces in natural images, such as frequency noise, whereas structural evidence mainly appears in local topology, text boundaries, and layout relations. SAP-DSP captures traditional visual artifacts through a texture stream and models structured generation traces through a structure stream. It also uses a structural artifact prior to dynamically modulate the structure prompt. Finally, a structure-aware routing module adaptively fuses texture and structure features, thereby enhancing the unified detection performance. The main contributions of this paper are summarized as follows:

(\romannumeral1) We construct an MLLM-generated image detection benchmark from GPT Image2 and Nano Banana2. This benchmark covers texture-dominated, structure-dominated, and hybrid-dominated. It is designed to evaluate detector performance under the new challenges introduced by large-scale image generation models.

(\romannumeral2) We systematically evaluate representative detectors on both public benchmarks and the proposed benchmark. We further analyze false positives on real images and false negatives on generated images. The results reveal the generalization limitations and failure modes of existing detectors.

(\romannumeral3) We propose SAP-DSP as a strong baseline. It jointly models texture cues and structural cues through a structural artifact prior, dual-stream prompt learning, and structure-aware routing fusion, achieving more stable performance on the proposed benchmark.

\section{Related Work}

\subsection{AI-Generated Image Detection Benchmarks}
Detection benchmarks are important for evaluating the generalization ability of AI-generated image detectors. Wang et al.~\cite{ref10} constructed a dataset containing images generated by multiple CNN-based generators~\cite{ref11, ref12, ref13}, which was used to study detector generalization across different generation architectures. Zhu et al.~\cite{ref5} proposed GenImage, a million-scale benchmark covering multiple advanced generators and a wide range of natural image categories. It provides an important platform for cross-generator detection and robustness evaluation. Pellegrini et al.~\cite{ref14} proposed AI-GenBench, which evaluates detectors in chronological order to simulate the arrival of new generators. Li et al.~\cite{ref15} further built a robustness benchmark from real-world scenarios and studied the effects of scene generalization. Zhang et al.~\cite{ref16} proposed TextFake for text-rich image detection. It covers multiple languages and shows that existing detectors clearly degrade on text-dense images.

These benchmarks have advanced the development of AI-generated image detection. However, their data sources are still mainly based on traditional generative models, with limited coverage of MLLMs. Unlike previous studies, this paper focuses on images generated by current large-scale models and constructs a benchmark dataset by incorporating multiple generation paths.

\begin{figure*}[ht]
  \centering
  \includegraphics[width=\linewidth]{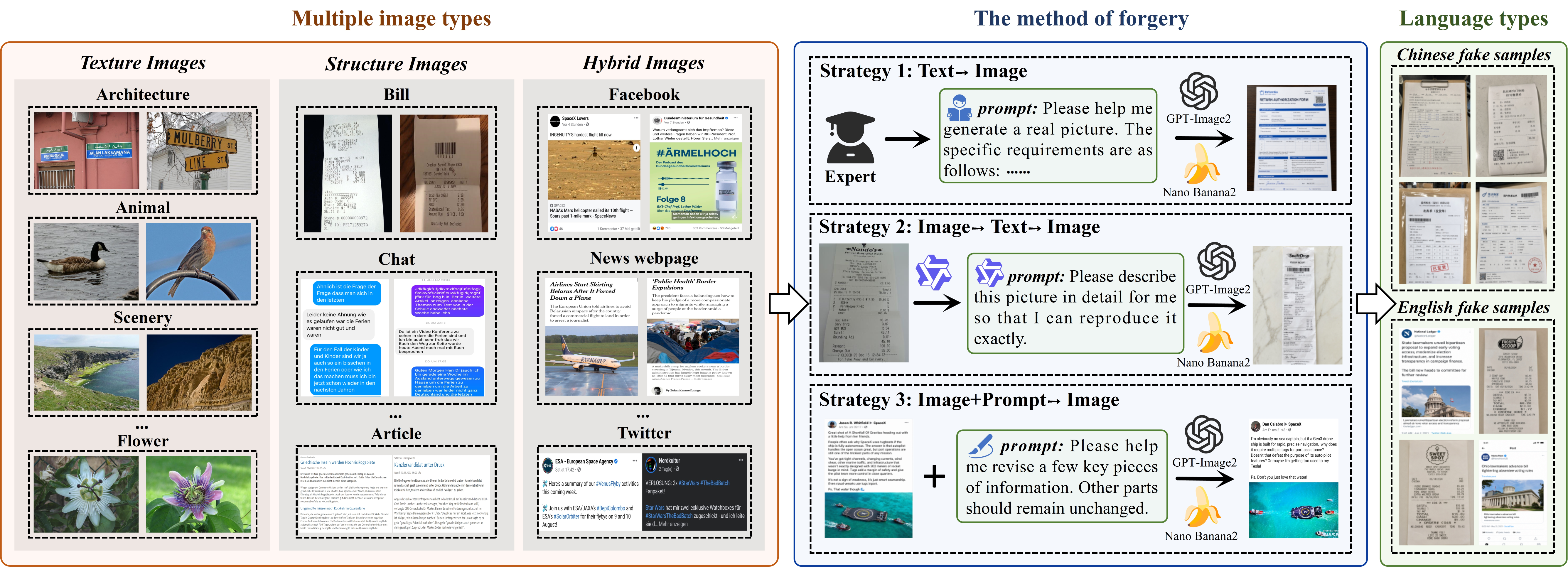}
  \caption{Overview of the dataset construction pipeline}
  \label{fig2}
\end{figure*}

\subsection{Detection Methods for AI-Generated Images}
AI-generated image detection aims to distinguish real images from images synthesized by generative models. It is an important task in multimedia forensics and visual security. Wang et al.~\cite{ref10} studied the general detection of CNN-generated images and found that classifiers trained with data augmentation can generalize to different generators. Ojha et al.~\cite{ref4} explored cross-generator detection with pretrained visual models, showing that large-scale pretrained representations are helpful for detecting images from unseen generators. For diffusion-generated images, Wang et al.~\cite{ref17} proposed DIRE, which uses diffusion reconstruction error to characterize the difference between real and generated images. Chen et al.~\cite{ref18} proposed DRCT, which improves detector generalization for diffusion-generated images through diffusion reconstruction and contrastive training.

Recently, vision-language representations have also been used for AI-generated image detection. Cozzolino et al.~\cite{ref19} built a lightweight detector based on CLIP features~\cite{ref20, ref21, ref22} and verified its generalization ability across generators and degradation scenarios. These methods have pushed AI-generated image detection from generator-specific settings toward more general detection. However, most existing methods still focus on natural images, and their decision cues mainly rely on texture noise and color anomalies. For documents and hybrid text-image content, generation traces are more closely related to local geometric structures. Therefore, it remains important to jointly model texture evidence and structural evidence.

\section{Benchmark Construction from MLLMs}
We construct a benchmark dataset to evaluate detector generalization in realistic large-model generation scenarios. Different from existing datasets that mainly contain faces, scenes, and other natural images, our dataset focuses on structured visual content and hybrid text-image content, while also retaining a certain proportion of natural images. This design forms a comprehensive benchmark that is closer to real-world applications.

\subsection{Dataset Overview}
The overall construction process of the dataset is shown in Figure~\ref{fig2}. The benchmark is built based on GPT Image2 and Nano Banana2. It is designed along three dimensions: visual content type, generation protocol, and language type. In terms of visual content, the dataset contains texture-dominated images, structure-dominated images, and hybrid content images, corresponding to natural texture scenes, regular structural scenes, and text-image mixed scenes, respectively. In terms of generation protocols, the dataset adopts \textit{text-to-image}, \textit{image-to-text-to-image}, and \textit{image+prompt-to-image} generation, which simulate different generation paths in real-world usage. In terms of language type, the dataset contains both Chinese and English samples, enabling evaluation under different layouts and text structures. The dataset can be denoted as:
\begin{equation}
  \mathcal{D}=\mathcal{D}_{tex} \cup \mathcal{D}_{str} \cup \mathcal{D}_{hyb}
\end{equation}

\subsection{Visual Content Types}
We divide the dataset into three content types according to their main visual attributes.

\textbf{Texture-dominated Images.}
These images mainly contain natural visual content, such as faces, animals, scenes, buildings, and indoor environments. Their visual information is characterized by natural textures, material details, smooth color variations, and realistic illumination. Generation traces in such images often appear as local texture anomalies, material inconsistencies, unnatural color transitions, abnormal high-frequency noise, or distorted details.

\textbf{Structure-dominated Images.}
These images mainly contain regular structures, such as document pages, receipts, tables, and web interfaces. Their visual information is characterized by text boundaries, line structures, table grids, and page layouts. Generation traces are often hidden in character strokes, edge information, line continuity, local alignment, and page-structure consistency, rather than appearing as obvious natural texture artifacts.

\textbf{Hybrid-dominated Images.}
These images contain both natural visual regions and structured layout regions. Examples include news pages, social media posts, web screenshots, and mixed image-text reports. Compared with purely natural images or purely structured images, hybrid content is closer to real-world online propagation and more challenging for detection. Detectors must consider both texture artifacts in natural regions and structural artifacts in text regions.

\subsection{Image Generation Protocols}
We design three protocols to simulate diverse forgery paths in real-world scenarios.

\textbf{Text-to-Image.}
This protocol directly generates an image from a text prompt:
\begin{equation}
  x_f = G_{\theta}(t)
\end{equation}
where $t$ denotes a manually designed text prompt, $G_{\theta}$ denotes the image generator, and $x_f$ denotes the generation of an image. This protocol simulates the generation of natural images, structured pages, or hybrid content from textual descriptions.

\textbf{Image-to-Text-to-Image.}
This protocol first converts a real image into a textual description and then regenerates an image from that description:
\begin{equation}
  t = V_{\phi}(x_r), \quad x_f = G_{\theta}(t)
\end{equation}
where $x_r$ denotes a real image and $V_{\phi}$ denotes an image captioning model. This protocol simulates semantic reconstruction from a real sample. The generated image is usually closer to the real sample in both content and layout, making it more deceptive than pure text-to-image generation.

\begin{figure*}[ht]
  \centering
  \includegraphics[width=\linewidth]{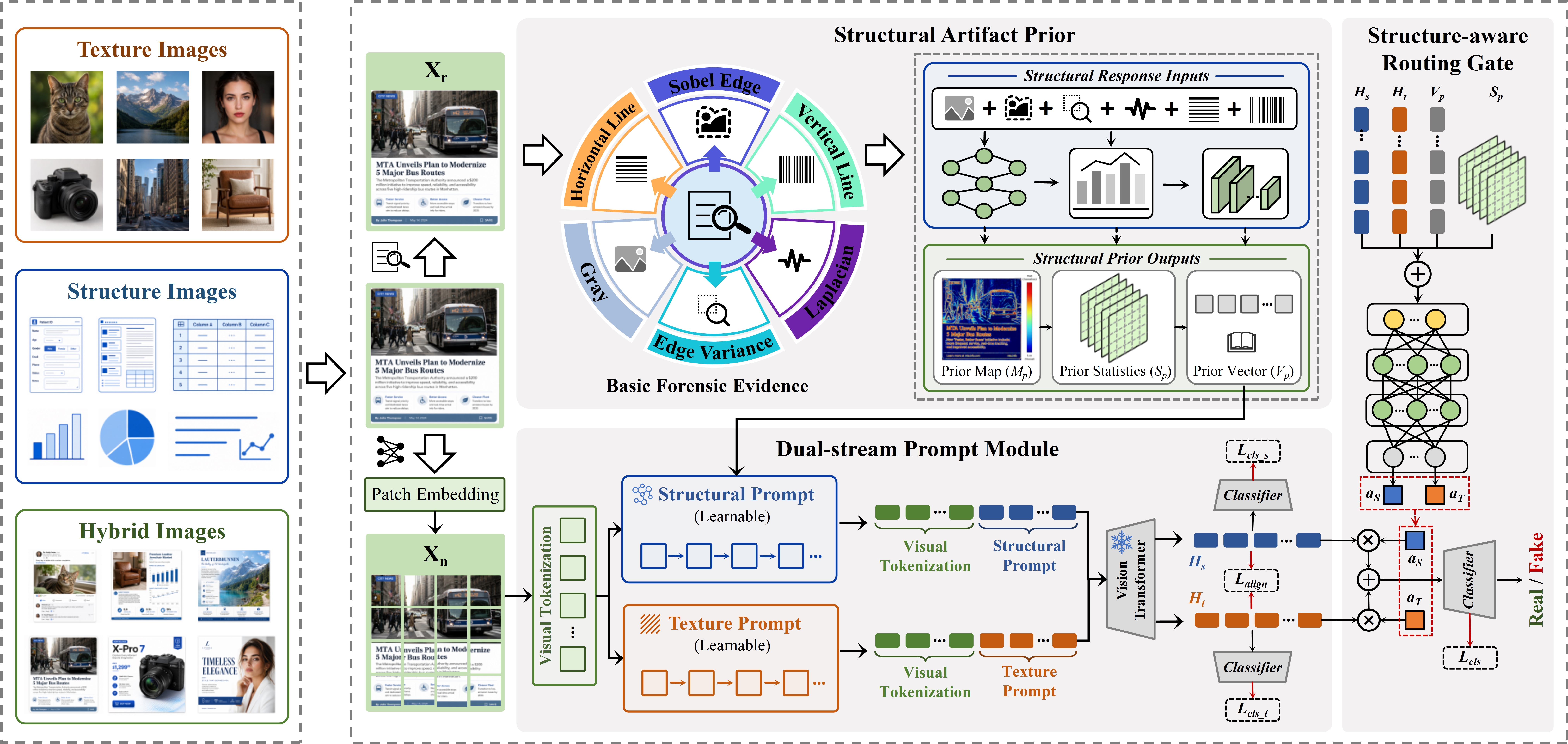}
  \caption{Overall architecture of the proposed SAP-DSP framework}
  \label{fig3}
\end{figure*}

\textbf{Image+Prompt-to-Image.}
This protocol uses a real image and an editing prompt as input, generating a modified image while keeping the main structure relatively stable:
\begin{equation}
  x_f = G_{\theta}(x_r, t_e)
\end{equation}
where $t_e$ denotes the editing prompt. This protocol simulates local tampering, such as modifying text in a web screenshot, replacing local content in a social media page, or editing key information in receipts and documents. Because the traces are often localized, this protocol is highly challenging for detectors.

\section{Methodology}
Figure~\ref{fig3} shows the proposed SAP-DSP framework. The method first extracts structural responses from the original image, which are encoded as a structural prior. It then constructs a texture stream and a structure stream to model texture artifacts and structural artifacts, respectively. Finally, a structure-aware routing module adaptively fuses the features from the two streams for AI-generated image detection.

\subsection{Structural Artifact Prior Module}
The structural artifact prior (SAP) module explicitly extracts structured forensic cues from the pixel space. We construct a set of lightweight structural response operators to characterize basic structural evidence. Given the input image $x$, this paper constructs two forms of input: the original pixel image $x_r$ and the standardized image $x_n$. For the raw image $x_r$, we first convert it into a grayscale map $G_r$:
\begin{equation}
  G_r = 0.299R + 0.587G + 0.114B
\end{equation}
where $R$, $G$, and $B$ denote the three color channels. Based on $G_r$, SAP extracts five structural responses: the Sobel edge response $R_e$, Laplacian high-frequency response $R_l$, horizontal line response $R_h$, vertical line response $R_v$, and local edge-variance response $R_{var}$. These responses characterize boundary structures, high-frequency variations, directional lines, and local structural stability. They can capture text-edge jitter, broken lines, table misalignment, abnormal UI borders, and local geometric discontinuities. The five types of responses are concatenated along the channel dimension to form the structural response tensor $R_s$:
\begin{equation}
  R_s = \operatorname{Concat}(R_e, R_l, R_h, R_v, R_{var})
\end{equation}

Subsequently, $R_s$ is fed into the lightweight prior head $f_p(\cdot)$ to obtain the structural artifact-aware response map $M_p$. It should be noted that $M_p$ is not a segmentation result obtained by direct supervision with pixel-level artifact annotations. Instead, it is a weakly supervised structure-aware map learned from structural responses. The prior head $f_p(\cdot)$ is trained end to end with the subsequent detection network, and no additional pixel-level annotations are required. $M_p$ indicates regions in the image that may contain structural artifacts or abnormal structural changes. Then, we further extract image-level structural statistics $S_p$ from $R_s$ and $M_p$:
\begin{equation}
\begin{aligned}
S_p = [&\mu(R_e), \mu(R_l), \mu(R_h), \mu(R_v), \\
      &\mu(R_{\mathrm{var}}), \mu(M_p), \sigma(M_p), \max(M_p)]
\end{aligned}
\end{equation}
Here, $\mu(\cdot)$, $\sigma(\cdot)$, and $\max(\cdot)$ denote the spatial mean, standard deviation, and maximum value, respectively. Subsequently, the structural statistics $S_p$ are encoded into the structural prior vector $V_p$ through the multi-layer perceptron $\operatorname{MLP_p}$:
\begin{equation}
  V_p = \operatorname{MLP_p}(S_p)
\end{equation}

Therefore, the SAP module forms a modeling chain with the following structure:
\begin{center}
$R_s \to M_p \to S_p \to V_p$
\end{center}

Among them, $M_p$ provides a spatial-level structural artifact-aware response, while $S_p$ describes the overall complexity and response intensity of the image structure. $V_p$ serves as a compact structural prior, which is used for subsequent structure prompt modulation and structure-aware routing fusion. It should be emphasized that SAP does not directly use traditional operators as the final discriminative features. Instead, it converts basic structural responses into structural priors that can guide prompt learning and routing fusion, thereby integrating explicit structural evidence into the pretrained visual representation learning process.

\subsection{Dual-Stream Prompt Learning Module}
We design a dual-stream prompt learning module to model texture evidence and structural evidence separately. The module consists of a texture stream and a structure stream, which share the same pretrained ViT backbone. Different prompts guide the model to focus on different forensic cues. Specifically, the texture stream captures traditional visual artifacts, while the structure stream models structured artifacts.

Let $P_t$ denote the texture prompt, which is a set of learnable prompt tokens. Since the texture stream does not use structural prior modulation, $P_t$ learns general visual artifacts from the image content. Let $P_s$ denote the structure prompt. It is also learnable, but is dynamically modulated by the structural prior vector $V_p$. This paper generates the scaling parameter $\alpha$ and the offset parameter $\beta$ from $V_p$ using a lightweight modulation network $\operatorname{MLP_m}$:
\begin{equation}
  [\alpha,\beta] = \operatorname{MLP_m}(V_p)
\end{equation}

Then, the structure prior modulation is applied to obtain the structure prompt $P'_s$:
\begin{equation}
  P'_s = P_s \odot (1 + \alpha) + \beta
\end{equation}
where $\odot$ denotes element-wise multiplication. This design allows the structure prompt to adapt to the structural attributes of each input image. For a normalized image $x_n$, the visual backbone first splits it into patch tokens. We insert the texture prompt $P_t$ and the structure prompt $P'_s$ at the input layer to form two prompt-conditioned inputs. The two streams share the same ViT encoder parameters, and the backbone remains frozen. Under different prompt conditions, the model obtains texture visual representation $h_t$ and structure visual representation $h_s$, which are then projected into a unified forensic feature space:
\begin{equation}
z_t = \operatorname{Norm}(g_t(h_t)),\quad
z_s = \operatorname{Norm}(g_s(h_s))
\end{equation}
where $z_t$ denotes the texture forensic feature, and $z_s$ denotes the structural forensic feature. $\operatorname{Norm}(\cdot)$ denotes $L_2$ normalization. To ensure that both streams have basic discriminative capabilities, we introduce auxiliary classification losses for the texture stream and the structural stream during training. The texture-stream loss $\mathcal{L}_{cls\_t}$ and the structure-stream loss $\mathcal{L}_{cls\_s}$ are computed as follows:
\begin{equation}
  \mathcal{L}_{cls\_t} = \mathcal{L}_{CE}(o_t, y), \quad
  \mathcal{L}_{cls\_s} = \mathcal{L}_{CE}(o_s, y)
\end{equation}
Here, $o_t$ and $o_s$ denote the auxiliary classification predictions of the texture stream and the structure stream, respectively. $y \in \{0,1\}$ denotes the real/generated class label, and $\mathcal{L}_{CE}$ denotes the cross-entropy loss function.

Using only classification supervision may lead to inconsistent feature spaces between the two streams, which can weaken the subsequent fusion stage. Therefore, we introduce a cross-stream alignment loss to align texture features $z_t$ and structure features $z_s$ at the class-prototype level. This avoids forcing each sample to have identical features in the two streams. In a mini-batch, for each class $c \in \{0,1\}$, the class prototypes of the texture stream and the structure stream are computed separately:
\begin{equation}
\mu_t^c=\frac{1}{|\mathcal{B}_c|}
\sum_{i\in\mathcal{B}_c} z_t^i,
\quad
\mu_s^c=\frac{1}{|\mathcal{B}_c|}
\sum_{i\in\mathcal{B}_c} z_s^i
\end{equation}
where $\mathcal{B}_c$ denotes the set of samples belonging to class $c$. The alignment loss $\mathcal{L}_{align}$ is defined as:
\begin{equation}
\mathcal{L}_{align}
=
\frac{1}{|C_B|}
\sum_{c \in C_B}
\left(
1-\operatorname{sim}(\mu_t^c,\mu_s^c)
\right)
\end{equation}
where $C_B$ denotes the set of classes appearing in the current mini-batch, and $\operatorname{sim}(\cdot,\cdot)$ denotes cosine similarity. This loss only aligns the overall prototypes of the texture stream and the structure stream within the same class, without forcing $z_t^i$ and $z_s^i$ of each sample to be exactly the same. Therefore, the two streams can remain consistent in the real/generated discrimination space while preserving their respective detailed differences and forensic focuses.
\begin{figure*}[ht]
  \centering
  \includegraphics[width=\linewidth]{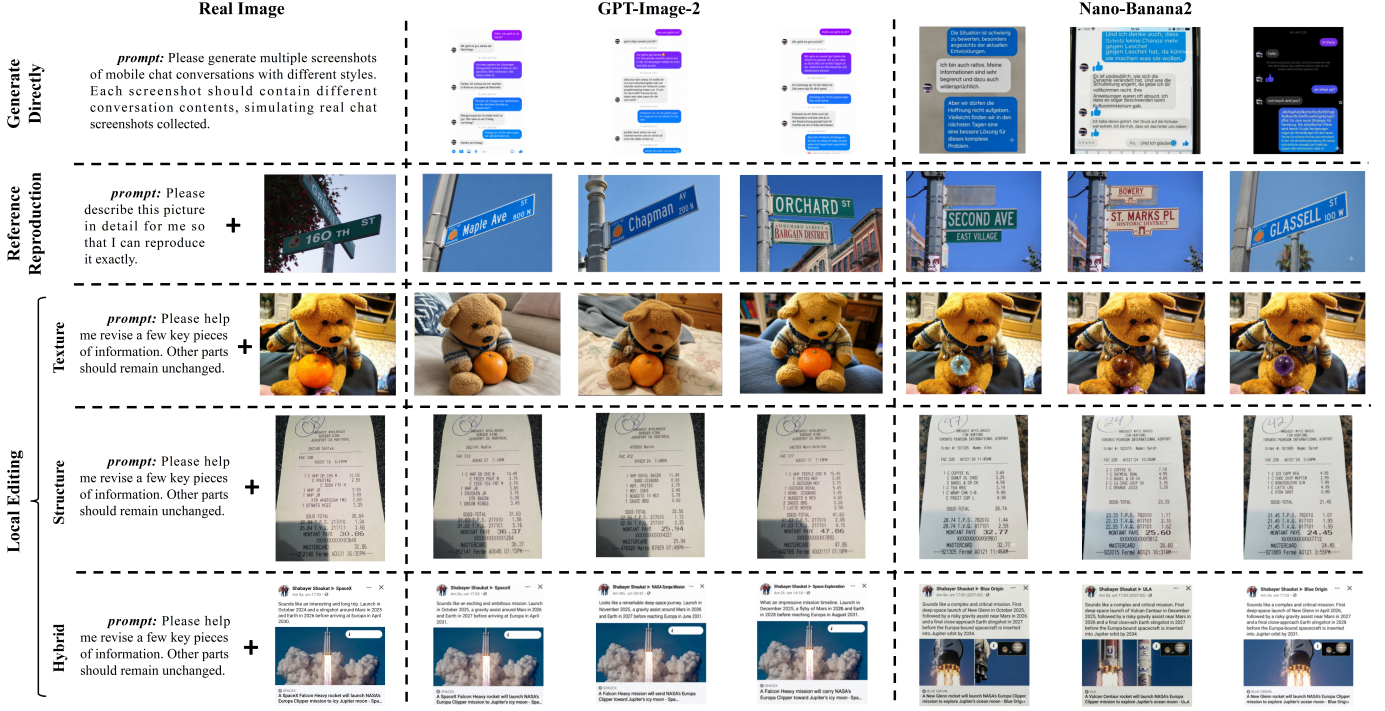}
  \caption{Examples of different data types in the benchmark}
  \label{fig4}
\end{figure*}

\subsection{Structure-Aware Routing Fusion Module}
Different image types rely on texture evidence and structural evidence to different degrees. Therefore, we design a structure-aware routing module to adaptively fuse the two streams. Unlike gated fusion based solely on feature concatenation, the structure-aware routing module explicitly incorporates the structural statistics $S_p$ and the structural prior vector $V_p$ into weight prediction. In this way, the fusion process is constrained by the complexity of the image structure and the intensity of structural artifacts. Specifically, the routing module takes the texture feature $z_t$, structural feature $z_s$, structural prior vector $V_p$, and structural statistics $S_p$ as inputs to predict the fusion weights of the two streams. The calculation is formulated as follows:
\begin{equation}
  [\alpha_t,\alpha_s] = \operatorname{Softmax}(\operatorname{MLP_r}([z_t; z_s; V_p; S_p]))
\end{equation}
where $\alpha_t$ and $\alpha_s$ denote the weights of the texture stream and the structure stream, respectively, satisfying $\alpha_t+\alpha_s=1$. The final fused feature $z_f$ is defined as:
\begin{equation}
  z_f = \operatorname{Norm}(\alpha_t z_t + \alpha_s z_s)
\end{equation}

By introducing structural statistics and structural prior vectors, the routing module performs fusion based not only on deep visual features, but also on the explicitly perceived structural properties of the input image. Then, the fused feature $z_f$ is fed into the main classifier to obtain the final binary classification prediction. The main classification loss $\mathcal{L}_{cls}$ is defined as:
\begin{equation}
  \mathcal{L}_{cls} = \mathcal{L}_{CE}(o, y)
\end{equation}
Here, $o$ denotes the output of the main classifier, and $y \in \{0,1\}$ denotes the real/generated label. By integrating the main classification loss, the bidirectional auxiliary classification losses, and the cross-stream alignment loss, the final training objective is defined as:
\begin{equation}
  \mathcal{L}_{total} = \mathcal{L}_{cls} + \lambda_t\mathcal{L}_{cls\_t} + \lambda_s\mathcal{L}_{cls\_s} + \lambda_a\mathcal{L}_{align}
\end{equation}
where $\lambda_t$, $\lambda_s$, and $\lambda_a$ control the auxiliary texture loss, auxiliary structure loss, and alignment loss, respectively.

\section{Experiment}
The experiments consist of two parts. First, we systematically evaluate the proposed benchmark by analyzing the generalization ability of existing detectors on images generated by MLLMs. Second, we compare SAP-DSP with existing methods and conduct ablation studies.

\subsection{Benchmark Evaluation}
We evaluate the benchmark from both macro and micro perspectives. At the macro level, we compare the performance gap between public benchmarks and our benchmark to verify the overall challenges introduced by images generated by MLLMs. At the micro level, we analyze false positive rates and false negative rates across different content types, revealing how existing methods fail on both real and generated images.

\subsubsection{Dataset Description}
The benchmark covers three visual domains: texture-dominated, structure-dominated, and hybrid-dominated. Texture-dominated images mainly include natural images, such as animals, scenes, and indoor environments. Structure-dominated images mainly include documents, tables, web pages, and screenshot pages. Hybrid content images contain both natural texture regions and structured layout regions, such as news pages, social media posts, posters, and mixed image-text pages. Table~\ref{tab1} summarizes the dataset, and Figure~\ref{fig4} shows examples of different data types.
\begin{table}[t]
\centering
\caption{Details of the proposed dataset}
\label{tab1}
\begin{tabularx}{0.47\textwidth}{>{\centering\arraybackslash}m{1.1cm}>{\centering\arraybackslash}X>{\centering\arraybackslash}X>{\centering\arraybackslash}m{1.4cm}>{\centering\arraybackslash}X>{\centering\arraybackslash}X}
\toprule
\textbf{Model}                                                                   & \textbf{Type} & \textbf{Texture} & \textbf{Structural} & \textbf{Hybrid} & \textbf{Total} \\ \midrule
\multirow{2}{*}{\textbf{\begin{tabular}[c]{@{}c@{}}GPT\\ Image2\end{tabular}}}   & Real          & 216              & 313                 & 197             & 726            \\
                                                                                 & Fake          & 216              & 313                 & 197             & 726            \\ \midrule
\multirow{2}{*}{\textbf{\begin{tabular}[c]{@{}c@{}}Nano\\ Banana2\end{tabular}}} & Real          & 216              & 313                 & 197             & 726            \\
                                                                                 & Fake          & 216              & 313                 & 197             & 726            \\ \bottomrule
\end{tabularx}
\end{table}

\begin{table*}[!h]
\centering
\caption{Performance comparison across datasets (\%)}
\label{tab2}
\begin{tabularx}{\textwidth}{>{\centering\arraybackslash}X>{\centering\arraybackslash}X>{\centering\arraybackslash}X>{\centering\arraybackslash}X>{\centering\arraybackslash}X>{\centering\arraybackslash}X>{\centering\arraybackslash}X>{\centering\arraybackslash}X>{\centering\arraybackslash}X}
\toprule
\multirow{3}{*}{\textbf{Datasets}}                     & \multicolumn{8}{c}{\textbf{Methods}}                                                                                            \\ \cmidrule(l){2-9} 
                                                       & \multicolumn{2}{c|}{\textbf{LTD}~\cite{ref23}}                                                                                                 & \multicolumn{2}{c|}{\textbf{PGC}~\cite{ref24}}                                                                                                 & \multicolumn{2}{c|}{\textbf{GAPL}~\cite{ref25}}                                                                                                & \multicolumn{2}{c}{\textbf{IAPL}~\cite{ref26}}                                                                            \\ \cmidrule(l){2-9} 
                                                       & \textbf{Accuracy}                                        & \multicolumn{1}{c|}{\textbf{Precision}}                                       & \textbf{Accuracy}                                        & \multicolumn{1}{c|}{\textbf{Precision}}                                       & \textbf{Accuracy}                                        & \multicolumn{1}{c|}{\textbf{Precision}}                                       & \textbf{Accuracy}                                        & \textbf{Precision}                                       \\ \midrule
GenImage                                               & 91.62                                                    & \multicolumn{1}{c|}{99.99}                                                    & 98.30                                                    & \multicolumn{1}{c|}{—}                                                        & 96.70                                                    & \multicolumn{1}{c|}{99.60}                                                    & 96.70                                                    & —                                                        \\ \midrule
UFD                                                    & 96.90                                                    & \multicolumn{1}{c|}{99.51}                                                    & 90.60                                                    & \multicolumn{1}{c|}{98.00}                                                    & 97.20                                                    & \multicolumn{1}{c|}{99.80}                                                    & 95.61                                                    & 99.32                                                    \\ \midrule
\begin{tabular}[c]{@{}c@{}}GPT\\ Image2\end{tabular}   & \begin{tabular}[c]{@{}c@{}}82.52\\ (\textbf{↓11.74})\end{tabular} & \multicolumn{1}{c|}{\begin{tabular}[c]{@{}c@{}}83.87\\ (\textbf{↓15.88})\end{tabular}} & \begin{tabular}[c]{@{}c@{}}57.21\\ (\textbf{↓37.24})\end{tabular} & \multicolumn{1}{c|}{\begin{tabular}[c]{@{}c@{}}64.36\\ (\textbf{↓33.64})\end{tabular}} & \begin{tabular}[c]{@{}c@{}}75.07\\ (\textbf{↓21.88})\end{tabular} & \multicolumn{1}{c|}{\begin{tabular}[c]{@{}c@{}}61.28\\ (\textbf{↓38.42})\end{tabular}} & \begin{tabular}[c]{@{}c@{}}70.37\\ (\textbf{↓25.79})\end{tabular} & \begin{tabular}[c]{@{}c@{}}69.30\\ (\textbf{↓30.02})\end{tabular} \\ \midrule
\begin{tabular}[c]{@{}c@{}}Nano\\ Banana2\end{tabular} & \begin{tabular}[c]{@{}c@{}}80.71\\ (\textbf{↓13.55})\end{tabular} & \multicolumn{1}{c|}{\begin{tabular}[c]{@{}c@{}}91.03\\ (\textbf{↓8.72})\end{tabular}}  & \begin{tabular}[c]{@{}c@{}}72.78\\ (\textbf{↓21.67})\end{tabular} & \multicolumn{1}{c|}{\begin{tabular}[c]{@{}c@{}}84.19\\ (\textbf{↓13.81})\end{tabular}} & \begin{tabular}[c]{@{}c@{}}81.14\\ (\textbf{↓15.81})\end{tabular} & \multicolumn{1}{c|}{\begin{tabular}[c]{@{}c@{}}83.17\\ (\textbf{↓16.53})\end{tabular}} & \begin{tabular}[c]{@{}c@{}}72.25\\ (\textbf{↓23.91})\end{tabular} & \begin{tabular}[c]{@{}c@{}}85.20\\ (\textbf{↓14.12})\end{tabular} \\ \bottomrule
\end{tabularx}
\end{table*}

\begin{figure*}[!h]
  \centering
  \includegraphics[width=\linewidth]{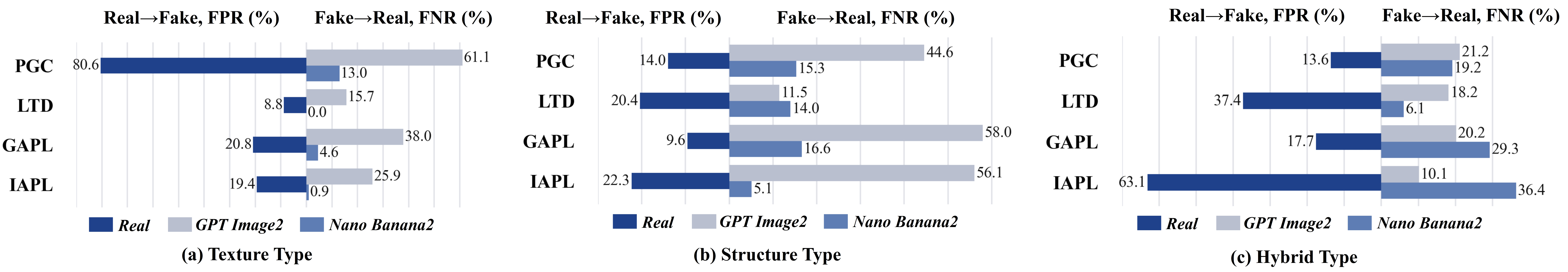}
  \caption{Error analysis under different content types}
  \label{fig5}
\end{figure*}

\subsubsection{Dataset Preprocessing}
To ensure consistency in input scale and file format across images from different sources, this paper adopts a unified preprocessing procedure for all real and generated images. First, image orientation is corrected based on EXIF information, and all images are uniformly converted to RGB format. Subsequently, the images are resized to a fixed resolution of $320\times320$ while preserving the integrity of the main content as much as possible. In the experimental setup, JPEG re-encoding and mild Gaussian blur are further introduced to approximately simulate secondary saving, platform compression, and slight visual degradation during online dissemination. Finally, all preprocessed images are saved in BMP format to avoid the influence of final file format differences on the detection results. This preprocessing procedure aims to reduce data bias caused by differences in image size, encoding method, and file format, and to evaluate the robustness of detection methods under mild propagation degradation conditions.

\subsubsection{Benchmark Transfer Gap Analysis}
To verify that the proposed benchmark differs from existing public datasets, we conduct a benchmark transfer gap analysis. We select four recent representative detectors, namely LTD, PGC, GAPL, and IAPL, and compare their performance on UFD~\cite{ref4}, GenImage~\cite{ref5}, and our benchmark. This experiment provides a cross-benchmark comparison and shows how detector performance changes when models are transferred from traditional public benchmarks to images generated by MLLMs.

As shown in Table~\ref{tab2}, existing detectors usually achieve high accuracy and precision on UFD and GenImage, indicating that they perform well under traditional AI-generated image detection settings. However, when the same detectors are transferred to our benchmark, their performance drops to varying degrees. This suggests that images generated by GPT Image2 and Nano Banana2 are closer to real-world applications in terms of realism and text rendering. As a result, the low-level artifacts and model-specific cues used by existing detectors become weaker and less stable. The transfer gap confirms the necessity of constructing a benchmark specifically designed for MLLMs.

\subsubsection{Error Analysis under Different Content Types}
Although the performance gap across benchmarks indicates that the overall dataset in this study is more challenging, it still cannot reveal exactly in which samples the existing detectors fail. Therefore, we further analyze errors from the perspective of content type. We divide the test samples into Texture, Structure, and Hybrid groups. For each group, we compute two types of errors for the four detectors. The false positive rate (FPR) is defined as the ratio of real images misclassified as fake, while the false negative rate (FNR) is defined as the ratio of generated images misclassified as real. For generated images, we further distinguish between GPT Image2 and Nano Banana2, which helps analyze the difficulty of different generation sources under different content types.

As shown in Figure~\ref{fig5}, different content types exhibit distinct error patterns. For Texture images, some methods show a high FPR on real images. For example, the Real-to-Fake FPR of PGC reaches 80.6\%, indicating that PGC easily classifies texture-complex or locally degraded real images as generated. PGC also has a high FNR of 61.1\% on GPT Image2 texture images, suggesting that such generated samples can weaken its decision cues. Overall, the four baseline methods show different degrees of instability across the three content types. The challenge of our benchmark comes not only from its higher overall difficulty, but also from the interaction among content type, generation source, and error type.

\begin{table*}[ht]
\centering
\caption{Detection results of different methods on the proposed benchmark (\%)}
\label{tab3}
\begin{tabularx}{\textwidth}{>{\centering\arraybackslash}X>{\centering\arraybackslash}m{3.5cm}|>{\centering\arraybackslash}X>{\centering\arraybackslash}X>{\centering\arraybackslash}X>
{\centering\arraybackslash}X>{\centering\arraybackslash}X>{\centering\arraybackslash}X>{\centering\arraybackslash}X>{\centering\arraybackslash}X>{\centering\arraybackslash}X}
\toprule
\multirow{3}{*}{\textbf{Datasets}}                                               & \multirow{3}{*}{\textbf{Methods}} & \multicolumn{9}{c}{\textbf{Data types}}                                                                                                                                                          \\ \cmidrule(l){3-11} 
                                                                                 &                                   & \multicolumn{3}{c|}{\textbf{Texture}}                                 & \multicolumn{3}{c|}{\textbf{Structure}}                               & \multicolumn{3}{c}{\textbf{Hybrid}}              \\ \cmidrule(l){3-11} 
                                                                                 &                                   & \textbf{Acc.}  & \textbf{Rec.}  & \multicolumn{1}{c|}{\textbf{F1}}    & \textbf{Acc.}  & \textbf{Rec.}  & \multicolumn{1}{c|}{\textbf{F1}}    & \textbf{Acc.}  & \textbf{Rec.}  & \textbf{F1}    \\ \midrule
\multirow{5}{*}{\textbf{\begin{tabular}[c]{@{}c@{}}GPT\\ Image2\end{tabular}}}   & GAPL~\cite{ref25}                       & 70.37          & 62.03          & \multicolumn{1}{c|}{67.67}          & 68.47          & 42.03          & \multicolumn{1}{c|}{57.14}          & 86.36          & 79.79          & 85.41          \\
                                                                                 & PGC~\cite{ref24}                        & 55.18          & 58.88          & \multicolumn{1}{c|}{57.56}          & 72.29          & 55.41          & \multicolumn{1}{c|}{66.67}          & 78.64          & 78.79          & 78.72          \\
                                                                                 & LTD~\cite{ref23}                        & 85.64          & 84.25          & \multicolumn{1}{c|}{85.44}          & 83.17          & \textbf{85.53} & \multicolumn{1}{c|}{84.13}          & 78.76          & 81.82          & 79.41          \\
                                                                                 & IAPL~\cite{ref26}                       & 77.77          & 74.07          & \multicolumn{1}{c|}{76.92}          & 67.19          & 43.94          & \multicolumn{1}{c|}{57.26}          & 66.16          & 89.89          & 72.65          \\
                                                                                 & \textbf{Ours}                     & \textbf{95.37} & \textbf{97.22} & \multicolumn{1}{c|}{\textbf{95.45}} & \textbf{87.89} & 80.89          & \multicolumn{1}{c|}{\textbf{86.98}} & \textbf{92.42} & \textbf{92.93} & \textbf{92.46} \\ \midrule
\multirow{5}{*}{\textbf{\begin{tabular}[c]{@{}c@{}}Nano\\ Banana2\end{tabular}}} & GAPL~\cite{ref25}                       & 87.51          & 95.37          & \multicolumn{1}{c|}{88.41}          & 84.71          & 83.44          & \multicolumn{1}{c|}{84.52}          & 71.21          & 70.71          & 71.07          \\
                                                                                 & PGC~\cite{ref24}                        & 47.22          & 87.04          & \multicolumn{1}{c|}{62.25}          & 83.76          & 84.71          & \multicolumn{1}{c|}{83.91}          & 87.37          & 80.81          & 86.49          \\
                                                                                 & LTD~\cite{ref23}                        & 92.69          & 93.16          & \multicolumn{1}{c|}{92.94}          & 77.71          & 85.99          & \multicolumn{1}{c|}{79.41}          & 71.72          & 93.94          & 76.86          \\
                                                                                 & IAPL~\cite{ref26}                       & 89.35          & 97.07          & \multicolumn{1}{c|}{90.30}          & 79.94          & \textbf{94.90} & \multicolumn{1}{c|}{82.55}          & 47.47          & 63.64          & 54.78          \\
                                                                                 & \textbf{Ours}                     & \textbf{96.07} & \textbf{97.32} & \multicolumn{1}{c|}{\textbf{96.88}} & \textbf{88.22} & 88.54          & \multicolumn{1}{c|}{\textbf{88.25}} & \textbf{92.93} & \textbf{96.97} & \textbf{93.20} \\ \bottomrule
\end{tabularx}
\end{table*}

\begin{table*}[!ht]
\centering
\caption{Ablation results under different datasets (\%)}
\label{tab4}
\begin{tabularx}{\textwidth}{>{\centering\arraybackslash}X>{\centering\arraybackslash}m{2.5cm}|>{\centering\arraybackslash}X>{\centering\arraybackslash}X>{\centering\arraybackslash}X>
{\centering\arraybackslash}X>{\centering\arraybackslash}X>{\centering\arraybackslash}X>{\centering\arraybackslash}X>{\centering\arraybackslash}X>{\centering\arraybackslash}X}
\toprule
\multirow{3}{*}{\textbf{Datasets}}                                               & \multirow{3}{*}{\textbf{Methods}} & \multicolumn{9}{c}{\textbf{Data types}}                                                                                                        \\ \cmidrule(l){3-11} 
                                                                                 &                                   & \multicolumn{3}{c|}{\textbf{Texture}}                            & \multicolumn{3}{c|}{\textbf{Structure}}                          & \multicolumn{3}{c}{\textbf{Hybrid}}         \\ \cmidrule(l){3-11} 
                                                                                 &                                   & \textbf{Acc.} & \textbf{Rec.} & \multicolumn{1}{c|}{\textbf{F1}} & \textbf{Acc.} & \textbf{Rec.} & \multicolumn{1}{c|}{\textbf{F1}} & \textbf{Acc.} & \textbf{Rec.} & \textbf{F1} \\ \midrule
\multirow{5}{*}{\textbf{\begin{tabular}[c]{@{}c@{}}GPT\\ Image2\end{tabular}}}   & SAP-DSP                           & 95.37         & 97.22         & \multicolumn{1}{c|}{95.45}       & 87.89         & 80.89         & \multicolumn{1}{c|}{86.98}       & 92.42         & 92.93         & 92.46       \\
                                                                                 & w/o SAP                           & 93.27         & 95.48         & \multicolumn{1}{c|}{94.77}       & 84.29         & 76.39         & \multicolumn{1}{c|}{81.27}       & 89.17         & 88.47         & 88.96       \\
                                                                                 & w/o TS                            & 90.37         & 92.57         & \multicolumn{1}{c|}{91.41}       & 86.66         & 78.67         & \multicolumn{1}{c|}{82.49}       & 88.79         & 87.49         & 87.93       \\
                                                                                 & w/o SS                            & 92.68         & 94.29         & \multicolumn{1}{c|}{93.66}       & 82.73         & 73.95         & \multicolumn{1}{c|}{77.54}       & 89.73         & 87.51         & 88.64       \\
                                                                                 & w/o SAR                           & 92.64         & 93.47         & \multicolumn{1}{c|}{92.97}       & 84.28         & 77.19         & \multicolumn{1}{c|}{80.92}       & 89.21         & 88.73         & 89.03       \\ \midrule
\multirow{5}{*}{\textbf{\begin{tabular}[c]{@{}c@{}}Nano\\ Banana2\end{tabular}}} & SAP-DSP                           & 96.07         & 97.32         & \multicolumn{1}{c|}{96.88}       & 88.22         & 88.54         & \multicolumn{1}{c|}{88.25}       & 92.93         & 96.97         & 93.20       \\
                                                                                 & w/o SAP                           & 93.71         & 94.83         & \multicolumn{1}{c|}{94.03}       & 84.19         & 82.78         & \multicolumn{1}{c|}{83.44}       & 89.34         & 93.88         & 91.21       \\
                                                                                 & w/o TS                            & 92.67         & 91.22         & \multicolumn{1}{c|}{91.39}       & 86.64         & 85.75         & \multicolumn{1}{c|}{86.08}       & 87.97         & 92.68         & 90.16       \\
                                                                                 & w/o SS                            & 92.84         & 93.83         & \multicolumn{1}{c|}{93.17}       & 82.43         & 80.68         & \multicolumn{1}{c|}{81.47}       & 88.15         & 92.28         & 90.61       \\
                                                                                 & w/o SAR                           & 93.07         & 94.81         & \multicolumn{1}{c|}{93.77}       & 84.91         & 86.28         & \multicolumn{1}{c|}{85.64}       & 88.84         & 93.18         & 90.51       \\ \bottomrule
\end{tabularx}
\end{table*}

\subsection{Method Evaluation}
This subsection compares SAP-DSP with advanced detectors and analyzes the main components of the proposed model.

\subsubsection{Experimental Settings}
To ensure a fair comparison, this paper adopts the same data partitioning strategy and evaluation protocol for SAP-DSP and all baseline methods. Specifically, we use a sample group-based training/testing partitioning strategy, where the same real image and its corresponding generated images are treated as one sample group. The same sample group is not allowed to appear in both the training and testing sets, thereby avoiding same-source sample leakage. The training and testing sets are divided at a ratio of 1:1, with approximately 50\% of the sample groups used for training and the remaining 50\% used for testing.

SAP-DSP is trained using the AdamW optimizer with a learning rate of 0.0003 and a batch size of 8. All input images are resized to a fixed resolution and normalized. The model is implemented based on the PyTorch framework and trained on an NVIDIA RTX 3090 GPU. We further evaluate the detection performance of the model on images generated by two large-scale models, GPT Image2 and Nano Banana2, and focus on analyzing its grouped performance on texture-dominated, structure-dominated, and hybrid-dominated images.

\subsubsection{Baseline Comparison}
We compare SAP-DSP with LTD, PGC, GAPL, and IAPL. Table~\ref{tab3} reports the detection results on the proposed benchmark. Existing methods achieve certain performance in AI-generated image detection. However, their overall performance is still affected by complex content types. This suggests that strategies mainly based on texture noise cannot fully handle structured visual content produced by current image generation models. In contrast, SAP-DSP achieves better overall results. Its advantage comes from two aspects. First, the texture stream captures texture noise, color anomalies, and local detail distortions, which are effective cues for natural images. Second, the structure stream focuses on structural regions, such as text boundaries, under the guidance of the structural artifact prior. The structure-aware routing module adaptively fuses texture evidence and structural evidence according to the structural attributes of the input image, thereby improving overall detection performance.

Some baseline methods perform relatively well on texture-dominated images. However, they clearly degrade on structure-dominated images. This indicates that generation traces in structured images differ from texture artifacts in natural images. Forgeries in structure-dominated images often appear in text edges and table lines, imposing higher requirements on detectors. SAP-DSP remains more stable across the three image types, indicating that the structural artifact prior and dual-stream prompt learning can mitigate complex structural problems in generated images.

\subsubsection{Ablation Study}
We conduct ablation experiments to verify the effectiveness of each component in SAP-DSP. The results are shown in Table~\ref{tab4}. The ablation settings include removing the structural artifact prior (SAP), removing the texture stream (TS), removing the structure stream (SS), and removing the structure-aware routing module (SAR).

Removing any core module leads to a performance drop, indicating that each module contributes to the final result. Without SAP, the structure stream lacks prior guidance from edges and high-frequency responses, weakening its ability to perceive key regions such as text boundaries. Without TS, the model mainly relies on structural cues and cannot effectively model texture noise and other artifacts in natural images. As a result, the performance on texture-dominated images drops significantly. Without SS, the model mainly relies on texture features and can still handle some texture-dominated images. However, its performance degrades more clearly on structure-dominated and hybrid images, showing that texture cues alone cannot fully describe structural artifacts. Without SAR, the model cannot dynamically adjust the contributions of the two streams and has to rely on fixed fusion, making it difficult to adapt to different content types.

\subsubsection{Evidence Visualization}
We further analyze the decision evidence of different detectors through visualization. The selected sample contains one real image and one generated image. The two images have highly similar backgrounds, object poses, and scene layouts, and differ mainly in a local semantic object region.

\begin{figure}[t]
  \centering
  \includegraphics[width=\linewidth]{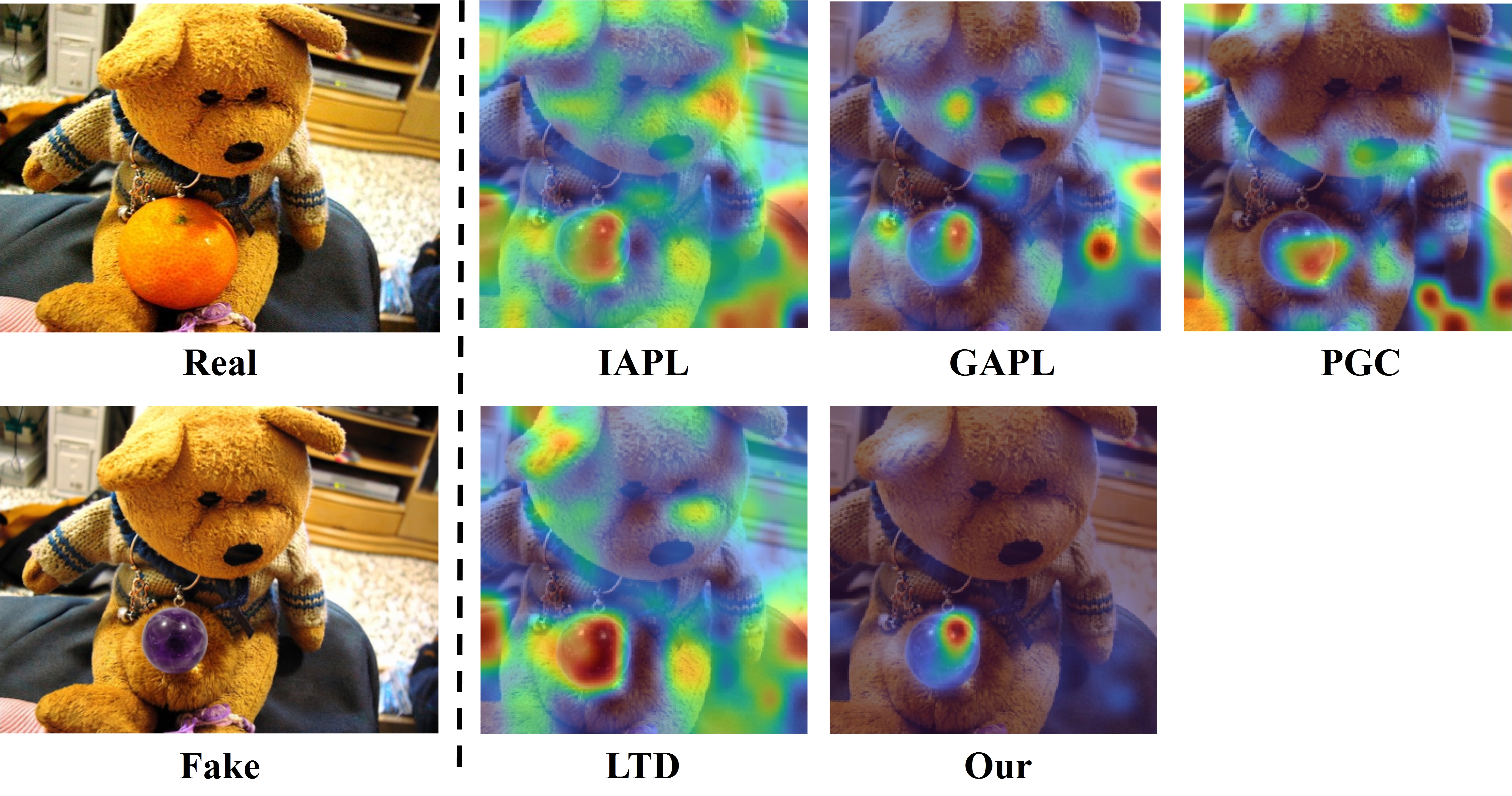}
  \caption{Visualization results of evidence from different methods}
  \label{fig6}
\end{figure}

Figure~\ref{fig6} shows the paired real and generated images on the left, while the right side presents the response heatmaps of advanced methods and SAP-DSP. The existing methods can generate responses in the image, but these responses are usually scattered and cannot focus on the core area of the forgery. In contrast, SAP-DSP produces more concentrated heatmap responses. The high-response area is mainly located near the local object region where the real and generated images differ. This result indicates that SAP-DSP does not rely solely on global color, background texture, or object semantics. Instead, it can capture local abnormal evidence introduced by generation or editing.

\section{Conclusion}
This paper constructs a benchmark for AI-generated image detection under MLLMs, including GPT Image2 and Nano Banana2. The benchmark covers texture-dominated, structure-dominated, and hybrid text-image application scenarios. Through systematic evaluation, we find that the strong performance of existing detectors on traditional public benchmarks does not directly transfer to the proposed benchmark. Existing methods exhibit clear failures in terms of both false positives on real images and false negatives on generated images. This verifies the detection challenges introduced by images generated by MLLMs. Based on this benchmark, we propose SAP-DSP as a strong structure-aware baseline. It improves the modeling of complex generation traces through a structural artifact prior, dual-stream prompt learning, and structure-aware routing fusion. Experiments show that SAP-DSP achieves more stable detection performance on the proposed benchmark. The results demonstrate the importance of structure-aware modeling for high-quality AI-generated image forensics.

\bibliographystyle{IEEEtran}
\bibliography{references}

\end{document}